%% file: main.tex
\documentclass[twoside,leqno,twocolumn]{article}

\usepackage[letterpaper]{geometry}

\usepackage{ltexpprt}

\usepackage{cite}
\usepackage{hyperref}
\hypersetup{colorlinks, citecolor=blue}

\input{mathdef}
\usepackage{amsmath,amssymb,amsfonts,mathtools}

\usepackage{graphicx}
\usepackage{subfigure}
\usepackage{float}
\usepackage{booktabs}
\usepackage{lscape}
\usepackage{bbm}

\usepackage{setspace}

\usepackage{epsfig,bm,subfigure,graphicx,color}
\usepackage{subfigure}
\usepackage{wrapfig}

\usepackage{algorithm, algorithmic}

\usepackage[version=3]{mhchem}

\usepackage{multirow}

\usepackage[font=small,skip=0.2cm]{caption}

\def\newprooftn#1#2{\newproof{#1}{Proof of Theorem #2}}
\def\newproofln#1#2{\newproof{#1}{Proof of Lemma #2}}

\begin{document}

\title{\Large Random Features Strengthen Graph Neural Networks}
\author{Ryoma Sato\thanks{Kyoto University, RIKEN AIP. \texttt{\{r.sato@ml.ist.i, myamada@i, kashima@i\}.kyoto-u.ac.jp}}
\and Makoto Yamada$^*$
\and Hisashi Kashima$^*$}

\date{}

\maketitle


\fancyfoot[R]{\scriptsize{Copyright \textcopyright\ 2021 by SIAM\\
Unauthorized reproduction of this article is prohibited}}





\begin{abstract} \small\baselineskip=9pt
Graph neural networks (GNNs) are powerful machine learning models for various graph learning tasks. Recently, the limitations of the expressive power of various GNN models have been revealed. For example, GNNs cannot distinguish some non-isomorphic graphs and they cannot learn efficient graph algorithms. In this paper, we demonstrate that GNNs become powerful just by adding a random feature to each node. We prove that the random features enable GNNs to learn almost optimal polynomial-time approximation algorithms for the minimum dominating set problem and maximum matching problem in terms of approximation ratios. The main advantage of our method is that it can be combined with off-the-shelf GNN models with slight modifications. Through experiments, we show that the addition of random features enables GNNs to solve various problems that normal GNNs, including the graph convolutional networks (GCNs) and graph isomorphism networks (GINs), cannot solve.
\end{abstract}

\section{Introduction}
\label{sec: introduction}

Graph neural networks (GNNs) ~\cite{gori, scarselli} have achieved state-of-the-art performance in various graph learning tasks, including chemo-informatics \cite{MPNNs}, question answering systems \cite{RCGN}, and recommender systems \cite{PinSAGE,KGCN,KGAT}. Recently, the theoretical power of GNNs have been extensively studied. Morris et al. \cite{kGNN} and Xu et al. \cite{GIN} pointed out that the expressive power of GNNs is at most the same as the $1$-dimensional Weisfeiler-Lehman (WL) test \cite{WLtest}. Sato et al. \cite{CPNGNN} considered the theoretical power of message-passing GNNs for combinatorial problems and demonstrated that the representation power of GNNs is the same as that of distributed local algorithms \cite{LocalSurvey}, and derived the approximation ratios of the algorithms that can be learned by GNNs. However, their approximation ratios are much higher than those of existing algorithms \cite{chlebik2008approximation, johnson1974approximation}. They proposed the use of feature engineering to improve these ratios; however, the improved ratios were still far from optimal.

In this paper, we propose a very simple and efficient method to improve the approximation ratios of GNNs, which can achieve near-optimal ratios under the degree-bounded assumption. Namely, we propose the addition of a random feature to each node. An illustrative example is shown in Figure \ref{fig: illust}. Message-passing GNNs cannot distinguish a node in a ring of three or six nodes if the node features are identical (Figure \ref{fig: illust} (a)). In contrast, if each node has a random feature, GNNs can determine the existence of a cycle of length three by checking whether there exists the same value as the root node at depth three (Figure \ref{fig: illust} (b)). Although this heuristic seems to work well and similar techniques such as relational pooling \cite{RelationalPooling} suggests its effectiveness, it is not trivial whether adding random features can improve the approximation ratios. In this paper, we propose graph isomorphism networks (GINs) with random features (rGINs), which add a random value to each node each time the procedure is called. We prove that the addition of random features indeed improves the theoretical capability of GNNs in terms of the approximation ratios. In a nutshell, our proposed method enables GNNs to learn randomized algorithms while standard GNNs learn only deterministic algorithms. Since it is known that distributed randomized algorithms are more powerful than distributed deterministic algorithms \cite{LocalSurvey}, our approach is expected to make GNNs more powerful. Table \ref{table: summary} summarizes our main results. Importantly, our results share a preferable characteristic with CPNGNNs \cite{CPNGNN}, which can be applied to graphs of variable sizes. We prove that there exist parameters such that the output of rGINs is not far from the optimal solution w.h.p. for any graph of \emph{arbitrary size}. This is the key difference from most of the previous works \cite{maron2019universality, keriven2019universal, dasoulas2019coloring}, including the relational pooling \cite{RelationalPooling}, where the upper bound of the graph size was fixed beforehand. Although the authors \cite{RelationalPooling} proposed to use a fixed constant support as ours, the effect of reducing the support is not clear, while we prove that GNNs are still powerful even if the support has only a constant number of elements. Thus this study provides an important step to theoretical results of GNNs for graphs of unbounded sizes and strengthens existing results.

In this study, we derive the approximation ratios of the algorithms that rGINs can learn by converting a certain type of constant time algorithms \cite{nguyen2008constant, rubinfeld2011sublinear} to rGINs. Conversely, we also prove that rGINs can be converted to constant time algorithms. This indicates that the advancement of GNN theory promotes the theory of constant time algorithms. Code and appendices are available at \url{https://github.com/joisino/random-features}.

\begin{table*}[tb]
\vspace{-0.2in}
\small
    \caption{The approximation ratios of the minimum dominating set problem (MDS) and maximum matching problem (MM). $^*$ indicates that these ratios match the lower bounds. 
    $\varepsilon > 0$ is an arbitrary constant, and $C$ is a fixed constant. The approximation ratios of rGINs match the best approximation ratios of polynomial algorithms except constant terms, and they also match the lower bounds except insignificant terms.} 
    \centering
    \scalebox{0.9}{
    \begin{tabular}{lccccc} \toprule
    \multirow{2}{*}{Problem} & GINs & CPNGNNs & \multirow{2}{*}{rGINs} & \multirow{2}{*}{Polynomial Time} & \multirow{2}{*}{Lower Bound} \\
    & CPNGNNs & + weak 2-coloring &  & & \\ \midrule
    \multirow{2}{*}{MDS} & $\Delta + 1^*$ & $\frac{\Delta + 1}{2}^*$ & $H(\Delta + 1) + \varepsilon$ & $H(\Delta + 1) - \frac{1}{2}$ & $\ln(\Delta) - C \ln \ln \Delta$ \\
    & \cite{CPNGNN} & \cite{CPNGNN} & (\textbf{This work}) & \cite{duh1997approximation} & \cite{chlebik2008approximation} \rule[-2mm]{0mm}{2mm} \\
    \multirow{2}{*}{MM} & $\infty^*$ & $\frac{\Delta + 1}{2}^*$ & $1 + \varepsilon^*$ & $1^*$ & \multirow{2}{*}{$1$} \\ 
    & \cite{CPNGNN} & \cite{CPNGNN} & (\textbf{This work}) & \cite{edmonds1965paths} & \\ \bottomrule
    \end{tabular}
    }
    \label{table: summary}
    \vspace{-.2in}
\end{table*}

\vspace{-.1in}
\section{Related Work}
The origin of GNNs dates back to Sperduti et al. \cite{sperduti1997supervised} and Baskin et al. \cite{baskin1997neural}, who aimed to extract features from graph data using neural networks. Gori et al. \cite{gori} and Scarselli et al. \cite{scarselli} proposed novel graph learning models that used recursive aggregation operations until convergence, and these models were called graph neural networks. Bruna et al. \cite{bruna2013spectral} and Defferrard et al. \cite{ChebyNet} utilized the graph spectral analysis and graph signal processing \cite{shuman2013emerging} to construct GNN models. Graph convolutional networks (GCNs) \cite{GCN} approximate a spectral model using linear filters to reduce it to an efficient spatial model. Gilmer et al. \cite{MPNNs} characterized GNNs using the message passing mechanism to provide a unified view of GNNs.

Although GNNs have been empirically successful, their limitations have been recently found. Morris et al. \cite{kGNN} and Xu et al. \cite{GIN} pointed out that the expressive power of GNNs is at most the same as the $1$-WL test \cite{WLtest}, and they cannot solve the graph isomorphism problem. \cite{maron2019provably} proposed a second-order tensor GNN model that has the same power as the $3$-WL test.
Relational pooling \cite{RelationalPooling} utilized all permutations of the nodes, similar to Janossy Pooling \cite{JanossyPooling}, to construct universal invariant and equivariant networks, and they proposed approximation schemes to make the computation tractable.
Sato et al. \cite{CPNGNN} showed that the representation power of GNNs is the same as that of distributed local algorithms \cite{angluin1980local, WeakModel, LocalSurvey}, which have the same representation power as model logic \cite{WeakModel}. Loukas \cite{loukas2019graph} and Barcel{\'{o}} et al. \cite{LogicalExpressiveness} demonstrated a similar connection between GNNs and distributed local algorithms and modal logic. Loukas \cite{loukas2019graph} characterized what GNNs cannot learn. In particular, he showed that message passing GNNs cannot solve many tasks even with powerful mechanisms unless the product of their depth and width depends polynomially on the number of nodes. In contrast, our main motivation is to show positive results of the expressive power. Although he also showed positive results, the assumptions (e.g., the Turing universality and unique node label) are much stronger than ours. We use GINs instead of a Turing universal model and use i.i.d. sampling to handle graphs with variable sizes.
After we finished this work, we noticed a parallel work \cite{dasoulas2019coloring} that showed adding coloring improved GNNs. The differences of their work and ours are two folds. First, they cannot handle graphs of variable sizes because they use permutations, while we can handle graphs of variable sizes because we use i.i.d. samplings. Second, we examined the expressive power of GNNs via the lens of approximation ratios and derived approximation ratios of graph algorithms that GNNs can learn, which cannot be derived from their results. A survey on the expressive power of GNNs is available in \cite{sato2020survey}

\vspace{0.1in} \noindent \textbf{RP-GNNs.} The proposed method is similar to $\pi$-SGD \cite{RelationalPooling}, one of approximation schemes for relational pooling, but they are different in two aspects: Our proposed method can be applied to graphs of variable sizes, whereas the original $\pi$-SGD cannot. This is because relational pooling uses a random permutation of $n$ elements, while we use i.i.d. random variables of a constant support. Although the authors proposed to use a fixed constant support as ours, the effect of reducing the support is not clear, while we prove that GNNs are still powerful even if the support has only a constant number of elements. Moreover, the original $\pi$-SGD aims to approximate an equivariant relational pooling layer, while \textbf{we aim to model non-equivariant functions using GNNs}. Note that our analysis assesses the power of GNNs via the lens of approximation ratios unlike relational pooling and provides another justification of $\pi$-SGD approximation.

\section{Background and Notations}
For a positive integer $k$, let $[k]$ be the set $\{1, 2, \dots, k\}$. Let $H(k) = \sum_{i = 1}^k \frac{1}{i}$ be the $k$-th harmonic number. Let $G = (V, E)$ be an input graph, where $V$ is a set of nodes, and $E$ is a set of edges. We only consider the connected graphs without self loops or multiple edges (i.e., connected simple graphs). $n = |V|$ denotes the number of nodes and $m = |E|$ denotes the number of edges. We assume $V = [n]$ without loss of generality. Let $V(G)$ be the set of the nodes of $G$ and $E(G)$ be the set of the edges of $G$. For a node $v \in V$, $\textrm{deg}(u)$ denotes the degree of node $v$, $\mathcal{N}_k(v)$ denotes the set of nodes within $k$-hop from node $v$, and $\mathcal{N}(v)$ denotes the set of neighboring nodes of node $v$. Let $R(G, v, L) = (\mathcal{N}_L(v), \{ \{x, y\} \in E \mid x, y \in \mathcal{N}_L(v) \})$ be the induced subgraph of $G$ by the $L$-hop nodes from node $v$. In some problem settings, each node of the input graph has a feature vector $\boldx_v \in \mathcal{C} \subset \mathbb{R}^{d_I}$. In such a case, we include feature vectors into the input graph $G = (V, E, \boldX)$, where $\boldX = [\boldx_1, \boldx_2, \ldots, \boldx_n]^\top \in \mathbbR^{n\times d_I}$ is the matrix for feature vectors.  We assume that the support $\mathcal{C}$ of the feature vectors is finite (i.e., $|\mathcal{C}| < \infty$). If the input graph involves no features, we use degree features as the initial embedding, following \cite{GIN, CPNGNN}.

\noindent \textbf{Assumption (Bounded-Degree Graphs).} In this paper, we only consider bounded-degree graphs, following \cite{CPNGNN}. There are many degree-bounded graphs in real world, such as chemical compounds and computer networks. Furthermore, the bounded-degree assumption is often used in constant time algorithms \cite{parnas2007approximating, nguyen2008constant}. It should be noted that this assumption is weaker than the bounded-size assumption because if the maximum degree $\Delta$ is equal to the maximum size of nodes, the bounded-degree graphs contain all bounded-size graphs. For each positive integer $\Delta \in \mathbb{Z}^+$, let $\mathcal{F} (\Delta)$ be the set of all simple connected graphs with maximum degrees of $\Delta$ at most. Let $\mathcal{F} (\Delta, \mathcal{C})$ be the set of all simple connected graphs $G = (V, E, \boldX)$ with maximum degrees of $\Delta$ at most with features $\boldx_v \in \calC$. 

\noindent \textbf{Definition (Node Problems).} A \textit{node problem} is a function $\Pi$ that associates a set $\Pi(G) \subseteq 2^V$ of \textit{feasible solutions} with each graph $G = (V, E)$.

\noindent \textbf{Definition (Edge Problems).} An \textit{edge problem} is a function $\Pi$ that associates a set $\Pi(G) \subseteq 2^E$ of \textit{feasible solutions} with each graph $G = (V, E)$.

We refer to the node and edge problems as graph problems. Many combinatorial graph problems aim to obtain a minimum or maximum set in feasible solutions. Let $\text{OPT}_m(\Pi, G)$ and $\text{OPT}_M(\Pi, G)$ denote the size of the minimum and maximum sets in $\Pi(G)$, respectively.

\noindent \textbf{Example.} A subset $U$ of $V$ is a dominating set if every vertex is in $U$ or is adjacent to at least one member of $U$. Let $\Pi_\text{MDS}(G)$ the set of all dominating sets in $G$. The minimum dominating set problem is the problem of computing $\text{OPT}_m(\Pi_\text{MDS}, G)$.

\noindent \textbf{Example.} A subset $F$ of $E$ is a matching if any two edges in $F$ are adjacent to each other. Let $\Pi_\text{MM}(G)$ be the set of all matchings in $G$. The minimum dominating set problem is the problem of computing $\text{OPT}_M(\Pi_\text{MM}, G)$.

These problems are fundamental in computer science with many applications in document summarization and resource allocation problems, among others.

\noindent \textbf{Definition (Monotonicity).} The minimization of a graph problem $\Pi$ is monotone if $\forall G = (V, E)$, $\forall S \subseteq T \subseteq V$, $S \in \Pi(G) \Rightarrow T \in \Pi(G)$. The maximization of a graph problem $\Pi$ is monotone if $\forall G = (V, E)$, $\forall S \subseteq T \subseteq V$, $T \in \Pi(G) \Rightarrow S \in \Pi(G)$.

Many combinatorial graph problems are monotone, such as the minimum dominating set problem, minimum vertex cover problem, and maximum matching problem. This property is used to ensure that the learned algorithm always outputs a feasible solution by including or excluding the uncertain nodes or edges.

\noindent \textbf{Definition (Consistent Algorithm).} An algorithm $\mathcal{A}$ that takes a graph $G = (V, E)$ as input and outputs a set of nodes or edges is a consistent algorithm of $\Pi$ if for all $G = (V, E)$, $\mathcal{A}(G) \in \Pi(G)$.

Note that algorithm $\mathcal{A}$ may involve randomized processes, but it must always output a feasible solution.

\noindent \textbf{Definition (Approximation Ratio).} The objective value $y$ of a minimization problem $\Pi$ is said to be an $(\alpha, \beta)$-approximation if $\text{OPT}_m(\Pi, G) \le y \le \alpha \text{OPT}_m(\Pi, G) + \beta$, and an objective value $y$ of a maximization problem $\Pi$ is an $(\alpha, \beta)$-approximation if $\frac{1}{\alpha} \text{OPT}_M(\Pi, G) - \beta \le y \le \text{OPT}_M(\Pi, G)$. The solution $S$ of a graph problem is also said to be an $(\alpha, \beta)$-approximation if $|S|$ is an $(\alpha, \beta)$-approximation. A consistent algorithm $\mathcal{A}$ is an $(\alpha, \beta)$-approximation algorithm for a graph problem $\Pi$ w.h.p. if for all graphs $G = (V, E)$, $\mathcal{A}(G)$ is an $(\alpha, \beta)$-approximation of $\Pi$ w.h.p. In particular, we refer to an $(\alpha, 0)$-approximation algorithm as an $\alpha$-approximation algorithm, and we call $\alpha$ the approximation ratio of the algorithm.

\begin{figure*}[tb]
\centering
\begin{minipage}{0.88\hsize}
\begin{minipage}{0.49\hsize}
\centering
\includegraphics[width=\hsize]{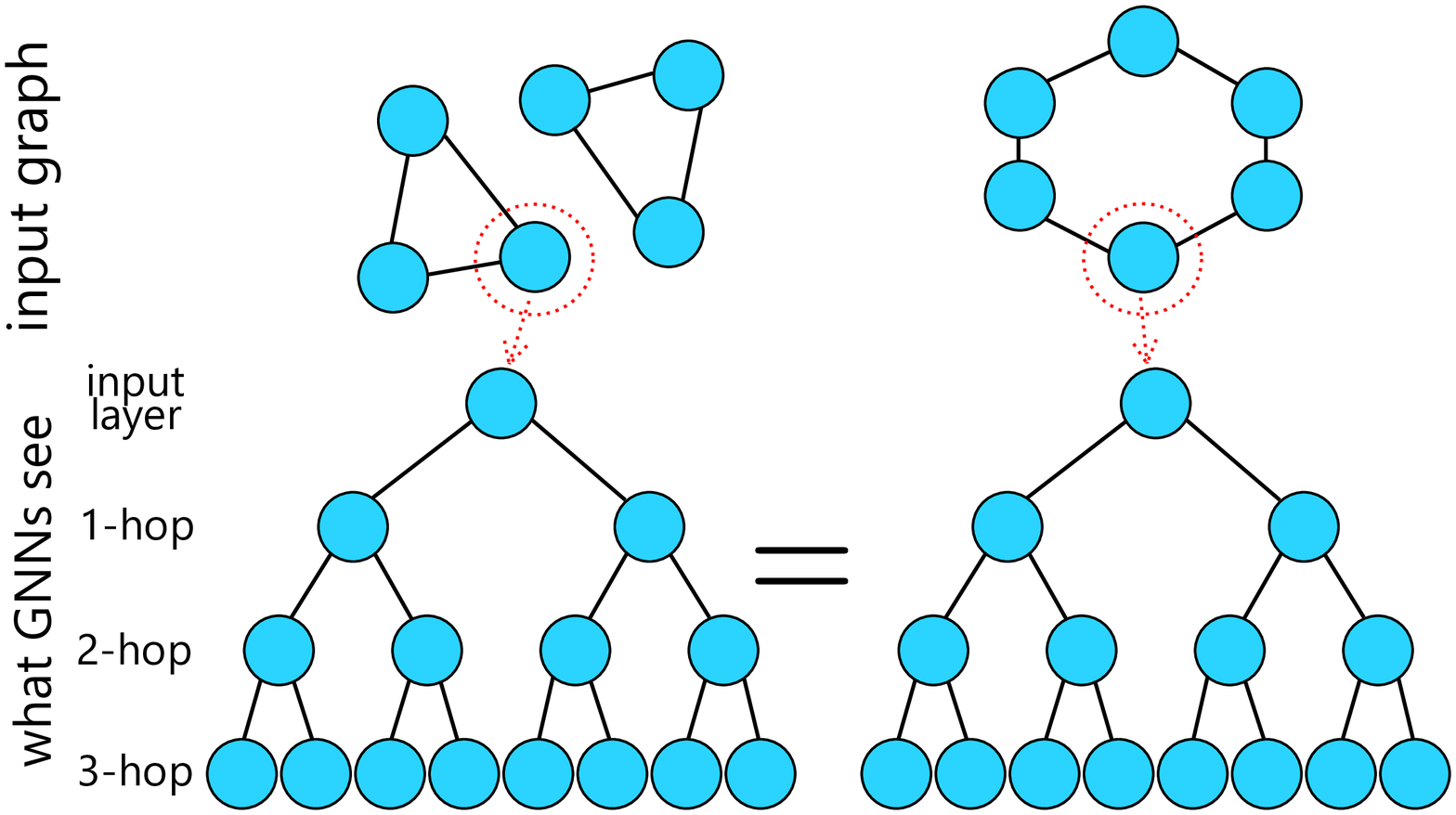}
(a) Identical Features.
\end{minipage}
\begin{minipage}{0.49\hsize}
\centering
\includegraphics[width=\hsize]{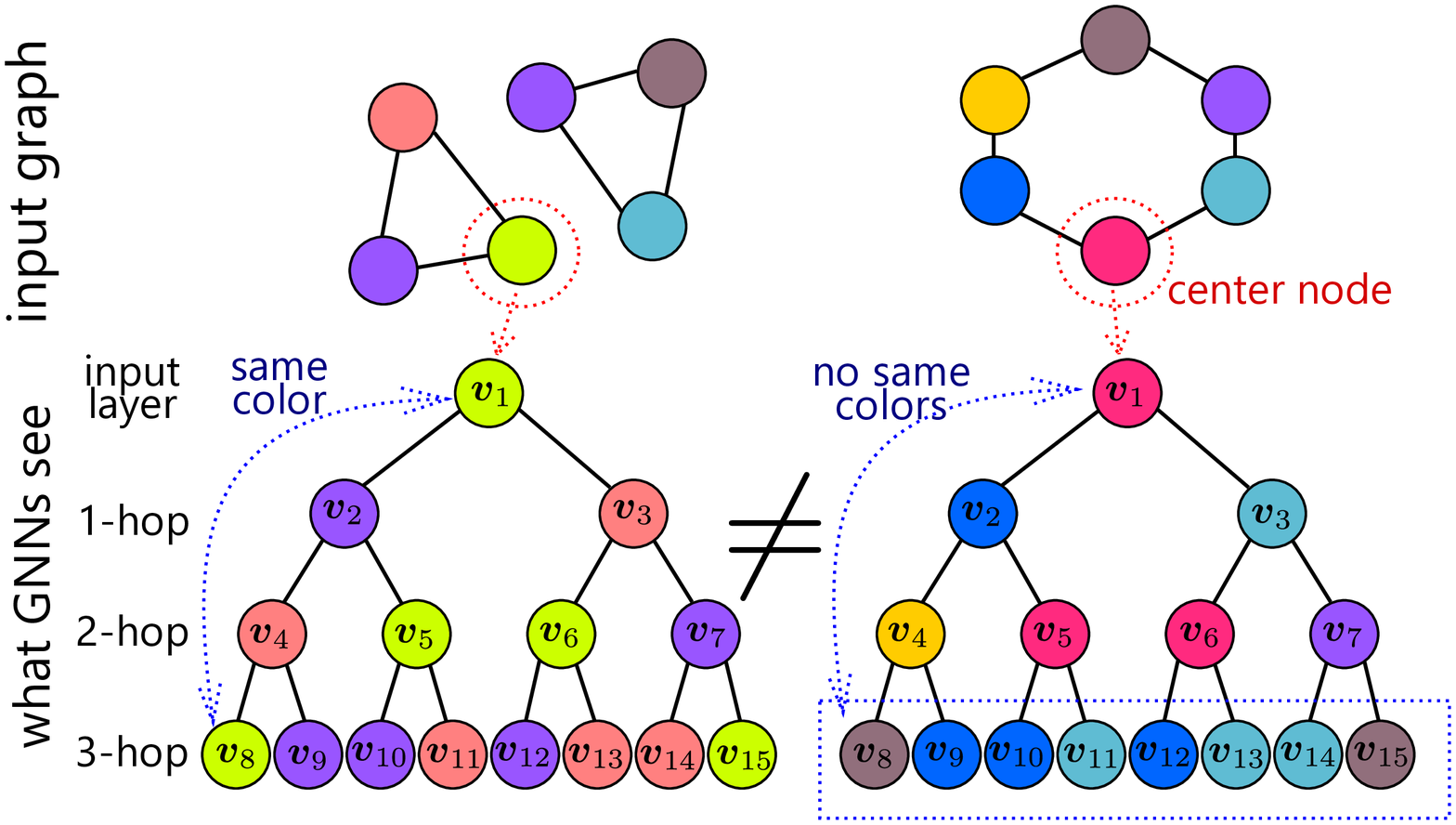}
(b) Random Features.
\end{minipage}
\end{minipage}
\vspace{-0.05in}
\caption{Illustrative example: GNNs with identical features (such as degree features) cannot distinguish a node in a cycle of three nodes with a node in a cycle of six nodes, whereas GNNs with random features can.}
\label{fig: illust}
\vspace{-0.2in}
\end{figure*}

\vspace{0.1in} \noindent \textbf{GINs.} Graph isomorphism networks (GINs) \cite{GIN} are a powerful model that takes a graph $G = (V, E, \boldX)$ as input and outputs an embedding $\boldz_v \in \mathbb{R}^{d_O}$ of each node $v \in V$. A GIN with parameters $\boldtheta$ calculates the embeddings $\boldz_v = \boldz^{(L)}_v$ by the following equations.
\begin{align*}
    \boldz^{(0)}_v &\leftarrow \text{MLP}_{\boldtheta_0}(\boldx_v), \\
    \boldz^{(l)}_v &\leftarrow \text{MLP}_{\boldtheta_l}\left((1 + \varepsilon^{(k)}) \boldz^{(l-1)}_v + \sum_{u \in \mathcal{N}(v)}\boldz^{(l-1)}_u\right),
\end{align*}
where $\text{MLP}_{\boldtheta_l}$ is a multi layer perceptron with parameters $\boldtheta_l$. Here, $\boldtheta$ includes $\boldtheta_0$, $\boldtheta_1$, $\dots$, $\boldtheta_L$ and $\varepsilon^{(1)}$, $\varepsilon^{(2)}$, $\dots$, $\varepsilon^{(L)}$. The existence of the parameters $\boldtheta$ implies that there exists an architecture (such as the number of layers and dimensions of hidden vectors) along with its parameters. We build a GNN model with random features based on GINs because GINs have the strongest power among message-passing GNNs \cite{GIN}. In particular, GINs can distinguish the neighboring node sets if the multisets of features of neighboring nodes are different and have the same power as $1$-WL.

\vspace{-.1in}
\section{Main Results} \label{sec: main}

\noindent \textbf{Intuitive Explanation.} We first provide an intuition using Figure \ref{fig: illust} (b). In this example, we use a toy model that concatenates features of all nodes and assume that the maximum degree is two for simplicity. The first dimension $\boldv_1$ of the embedding $\boldv$ is the random feature of the center node. The second and third dimensions are the random features of the one-hop nodes (e.g., in the sorted order) with appropriate zero paddings. The fourth to seventh dimensions are the random features of the two-hop nodes. The eighth to fifteenth dimensions are the random features of the three-hop nodes. Then, as Figure \ref{fig: illust} (b) shows, irrespective of the random features, the center node is involved in a $3$-cycle if and only if there exists a leaf node of the same color as the center node unless the random features accidentally coincide. This condition can be formulated as $\boldv_1 = \boldv_8$ or $\boldv_1 = \boldv_9$ or $\dots$ or $\boldv_1 = \boldv_{15}$. Therefore, we can check whether the center node is involved in a $3$-cycle by checking the embedding on the union of certain hyperplanes $\{\boldv \mid \boldv_1 = \boldv_8\} \cup \{\boldv \mid \boldv_1 = \boldv_9\} \cup \dots \cup \{\boldv \mid \boldv_1 = \boldv_{15}\}$. This property is valid even if the random features are re-assigned; a center node involved in a cycle of length three always falls on the union of these hyperplanes irrespective of the random features. A similar property is valid for substructures other than a $3$-cycle. Therefore, if the positive examples of a classification problem have characteristic substructures, the model can classify the nodes by checking the embedding on certain hyperplanes. It is noteworthy that the values of random features are not important because the values are random; however, the relationship between the values is important. Although we use GINs in the main results, according to the theory of GINs, this model does not lose information, and similar discussion can be applied by replacing the hyperplanes with curved surfaces. The aggregation functions of GINs are parameterized, and appropriate features are selected according to the downstream task. For example, if the downstream task relies only on the existence of a $3$-cycle, GINs learn to discard the second to seventh dimensions in the example above. These intuitions are formally stated in Theorem \ref{thm: capability}. Note that similar arguments are true in non-identical features. For example, consider a $3$-cycle of C, N, and O, and a $6$-cycle of C, N, O, C, N, and O, where C, N, and O are node features (e.g., atom type). GNNs cannot distinguish the C node in the 3-cycle and a C node in the 6-cycle even though topology these node belong to is different. Adding random features help distinguish them as we discussed in the indentical features case.

In this paper, we further show that the GINs with random features can solve the combinatorial problems with lower approximation ratios than the existing GNNs, where distinguishing the local structures is not sufficient. For example, if the input graph is a clique, all nodes are isomorphic. However, the empty set does not form a dominating set and the entire nodes contain too many nodes because the minimum dominating set contains only one node. We show that random features can help select the nodes appropriately. This type of mechanism is not required for ordinary node classification tasks, but important for combinatorial problems and cannibalization-aware recommendations \cite{gong2019exact}.

\setlength{\textfloatsep}{0pt}
\begin{algorithm}[tb]
\caption{rGINs: GINs with random features}
\label{algo: rGIN}
\begin{algorithmic}
\REQUIRE $G = (V, E, \boldX)$, Distribution $\mu$, Parameters $\boldtheta$. 
\ENSURE Embeddings $[\boldz_1, \boldz_2, \dots, \boldz_n]^\top \in \mathbb{R}^{n \times d_O}$
\STATE Assign random features $\boldr_v \sim \mu \quad (\forall v \in V)$
\STATE \textbf{return} $\text{GIN}_\boldtheta((V, E, \textsc{CONCAT}([\boldX, [\boldr_1, \dots, \boldr_n]^\top])))$.
\end{algorithmic}
\end{algorithm}

\vspace{0.1in} \noindent \textbf{rGINs.}
In this section, we introduce GINs with random features (rGINs). rGINs assign a random value $\boldr_v \in D \subseteq \mathbb{R}^{d_r}$ to each node $v$ every time the procedure is called and calculate the embeddings of a node using GINs, and let $\boldR = [\boldr_1, \boldr_2, \dots, \boldr_n]^\top \in \mathbb{R}^{n \times d_r}$. Here, $D$ is the support of random features. We show the pseudo code of rGINs in Algorithm \ref{algo: rGIN}, where $\textsc{CONCAT}([\boldX, \boldR]) \in \mathbb{R}^{n \times (d_I + d_r)}$ denotes concatenation along the feature dimension. We show that this slight modification theoretically strengthens the representation power of GINs.

In the following analysis, we use discrete random features to ensure consistency of the learned algorithm. Continuous random features cannot ensure consistency because the theoretical analysis of GINs assumes that features are countable \cite{GIN}. However, continuous random features can be used in practice. For consistency algorithms, the continuous features can be discretized. 

We then introduce the property that random features should satisfy. We prove in the following sections that the quality of solutions generated by rGINs can be guaranteed if the random distribution of random features are i.i.d. with the following property.

\noindent \textbf{Definition ($\mathcal{U}(p)$).} For $p \in \mathbb{R}^+$, a discrete probability measure $\mu$ with support $D \subseteq \mathbb{R}^{d_r}$ has the property $\mathcal{U}(p)$ or $\mu \in \mathcal{U}(p)$ if $\mu(x) \le p$ for all $x \in D$.

\noindent \textbf{Example.} For all $p \in \mathbb{R}^+$, the uniform distribution $\text{Unif}(D)$ on $D = [\text{ceil}(\frac{1}{p})]$ has the property $\mathcal{U}(p)$.

\vspace{0.1in} \noindent \textbf{How to solve node problems using rGINs.} We solve the node problems using node classification models. We first compute the embeddings $\boldz_1, \dots, \boldz_n \in \mathbb{R}$ using rGINs with output dimension $d_O = 1$ and the sigmoid activation in the last layer. For each node $v$ that has an $L$-hop node $u \in \mathcal{N}_L(v)$ with the same random feature (i.e., $\boldr_v = \boldr_u$), we fix the embedding $\boldz_v = 1$ and $\boldz_v = 0$ for the monotone minimization and maximization problem, respectively. This step ensures that the learned algorithm is consistent, but this step is optional and can be skipped for some applications. Finally, we decide the solution $U = \{v \in V \mid \boldz_v > 0.5\}$ by setting a threshold for the output probabilities. $\text{rGIN}_V(G, \mu, \boldtheta)$ denotes the function that takes a graph $G = (V, E, \boldX)$ as input and returns $U$ by the procedure above.

\vspace{0.1in} \noindent \textbf{How to solve edge problems using rGINs.} We solve the edge problems using link prediction models. We first compute the embeddings of each node using rGINs. For each node $v$ that has an $L$-hop node $u \in \mathcal{N}_L(v)$ with the same random feature (i.e., $\boldr_v = \boldr_u$), we fix the embedding $\boldz_v = \mathbf{1}$ and $\boldz_v = \mathbf{0}$ for the monotone minimization and maximization problem, respectively, where $\mathbf{1} \in \mathbb{R}^{d_O}$ and $\mathbf{0} \in \mathbb{R}^{d_O}$ are vectors of ones and zeros, respectively. Finally, we decide the solution $F = \{\{u, v\} \in E \mid \boldz_u^\top \boldz_v > 0.5\}$ by setting a threshold for the inner product of embeddings, following a standard method for the link prediction task \cite{liben2007link}. $\text{rGIN}_E(G, \mu, \boldtheta)$ denotes the function that takes a graph $G = (V, E, \boldX)$ as input and returns $F$ by the procedure above.

\vspace{0.1in} \noindent \textbf{Expressive Power of rGINs.}
In this section, we demonstrate the expressive power of rGINs. Especially, we prove that rGINs can distinguish any local structure w.h.p. To prove the theorem, we first define an isomorphism between the pairs of a graph and a node.

\noindent \textbf{Definition} For $G = (V, E, \boldX)$, $G' = (V', E', \boldX')$, $v \in V$ and $v' \in V'$, $(G, v)$ and $(G', v')$ are isomorphic if there exists a bijection $f\colon V \to V'$ s.t. $f(v) = f(v')$, $(x, y) \in E \Leftrightarrow (f(x), f(y)) \in E'$, and $\boldx_{x} = \boldx'_{f(x)} ~(\forall x \in V)$. $(G, v) \simeq (G', v')$ denotes $(G, v)$ and $(G', v')$ are isomorphic.

\begin{theorem} \label{thm: capability}
$\forall L \in \mathbb{Z}^+$, $\exists p \in \mathbb{R}^+$ s.t. $\forall \mathcal{C} ~(|\mathcal{C}| < \infty)$, $\forall \mathcal{G} \subseteq \{(G, v) \mid G \in \mathcal{F}(\Delta, C), v \in V(G)\}$, $\forall \mu \in \mathcal{U}(p)$, there exist parameters $\boldtheta$ s.t. $\forall G = (V, E, \boldX) \in \mathcal{F}(\Delta, \mathcal{C})$, $\forall v \in V$,
\vspace{-0.1in}
\begin{itemize}
    \item if $\exists (G', v') \in \mathcal{G}$ such that $(G', v') \simeq (R(G, v, L), v)$ holds, $\textup{rGIN}(G, \mu, \boldtheta)_v > 0.5$ holds w.h.p.
    \item if $\forall (G', v') \in \mathcal{G}, ~(G', v') \not \simeq (R(G, v, L), v)$ holds, $\textup{rGIN}(G, \mu, \boldtheta)_v < 0.5$ holds w.h.p.
\end{itemize}
\end{theorem}

All proofs have been provided in the supplementary material. For example, let $L = 2$ and $\mathcal{G}$ be a set of all pairs of a graph and a node $v$ with at least one triangle incident to $v$. Then Theorem \ref{thm: capability} shows that rGINs can classify the nodes by presence of the triangle structure, while GINs cannot determine the existence of a triangle in general. We confirm this fact by numerical experiments in Section \ref{sec: experiments}. Moreover, let $\mathcal{G}$ be a set of all graphs with certain chemical functional groups, then rGINs can classify nodes based on the functional groups that the node belongs to.

\vspace{0.1in} \noindent \textbf{Minimum Dominating Set Problems.}
In this section, we modify a constant time algorithm for the minimum dominating set problem $\Pi_{\text{MDS}}$ and prove that rGINs can simulate this algorithm. Nguyen et al. \cite{nguyen2008constant} converted a sequential greedy algorithm for the minimum dominating set problem \cite{johnson1974approximation, lovasz1975ratio} into a constant time algorithm. We use a slightly different version of the sequential greedy algorithm.
\vspace{-0.05in}
\begin{enumerate}
    \item Assign a random value $\boldr_v \sim \mu$ to each node $v \in V$.
    \item Add nodes $v$ into the solution if there is a node $u \in \mathcal{N}_2(v)$ with the same random value as $v$.
    \item Add a node that covers the most number of uncovered nodes into the solution until all nodes are covered. Ties are broken by the lexicographical order of $\boldr_v$.
\end{enumerate}
\vspace{-0.05in}
This is a consistent algorithm by its stop criterion. Besides, the approximation ratio is bounded.

\begin{lemma} \label{lem: MDS_sequential}
For all $\varepsilon > 0$, there exists $p \in \mathbb{R}^+$ such that for any distribution $\mu \in \mathcal{U}(p)$, the algorithm above is an $(H(\Delta + 1) + \varepsilon)$ approximation algorithm for the minimum dominating set problem.
\end{lemma}

We construct a constant time algorithm using this sequential algorithm by constructing oracles $\mathcal{O}_0, \mathcal{O}_1, \dots, \mathcal{O}_{\Delta + 1}$ as \cite{nguyen2008constant}. Intuitively, $\mathcal{O}_0(v)$ is the indicator function that decides whether node $v$ is included in the solution after the second step of the greedy algorithm, and $\mathcal{O}_k$ is the indicator function that decides whether node $v$ is included in the solution when no node covers more than $\Delta + 1 - k$ uncovered nodes in the third step of the algorithm.
\vspace{-0.05in}
\begin{itemize}
    \item $\mathcal{O}_0(v)$ returns $1$ if there is a node $u \in \mathcal{N}_2(v)$ with the same random value as $v$ and returns $0$ otherwise.
    \item $\mathcal{O}_k(v)$ returns $1$ if $\mathcal{O}_{k-1}(v) = 1$. Otherwise, it queries $\{\mathcal{O}_{k-1}(u) \mid u \in \mathcal{N}_2(v) \text{ and } \boldr_u > \boldr_v \}$ and $\{\mathcal{O}_{k}(u) \mid u \in \mathcal{N}_2(v) \text{ and } \boldr_u < \boldr_v \}$ to determine that the number of uncovered nodes covered by $v$. If $v$ covers $\Delta + 2 - k$ uncovered nodes, $\mathcal{O}_k(v)$ returns $1$ and returns $0$ otherwise, where $\boldr_u < \boldr_v$ is a lexicographical comparison.
\end{itemize}
\vspace{-0.05in}
$\mathcal{O}_{\Delta + 1}(v)$ decides whether $v$ is in the solution of the greedy algorithm. Irrespective of the size of the input graph, this oracle stops within a constant number of steps w.h.p. by the locality lemma \cite{nguyen2008constant}.

\begin{lemma} \label{lem: MDS_local}
$\forall \varepsilon > 0$, there exist $L \in \mathbb{Z}^+$, $p \in \mathbb{R}^+$, and a function $f$ that takes a graph and a node as input and outputs a binary value s.t. $\forall \mu \in \mathcal{U}(p)$, $G = (V, E) \in \mathcal{F}(\Delta)$, let $\boldr_v \sim \mu ~(\forall v \in V)$. Then $(G, v) \simeq (G', v') \Rightarrow f(G, v) = f(G', v')$, $F = \{v \in V \mid f(R((V, E, \boldR), v, L), v) = 1\}$ always forms a dominating set, $\{ v \in V \mid \exists s, t \in \mathcal{N}_L(v) \text{ s.t. } \boldr_s = \boldr_t \} \subseteq F$ always holds, and $|F| \le (H(\Delta + 1) + \varepsilon) ~\textup{OPT}_m(\Pi_{\text{MDS}}, G)$ w.h.p. 
\end{lemma}

It can be proved that rGINs can simulate the function $f$ above and can learn an approximation algorithm for the minimum dominating set problem with a small approximation ratio (Table \ref{table: summary}).

\begin{theorem} \label{thm: MDS}
$\forall \varepsilon > 0$, there exist parameters $\boldtheta$, $p \in \mathbb{R}^+$ such that for all distributions $\mu \in \mathcal{U}(p)$ and graphs $G \in \mathcal{F}(\Delta)$, $\textup{rGIN}_V(G, \mu, \boldtheta) \in \Pi_{\text{MDS}}(G)$ always holds and $|\textup{rGIN}_V(G, \mu,  \boldtheta)| \le (H(\Delta + 1) + \varepsilon) ~\textup{OPT}_m(\Pi_{\text{MDS}}, G)$ holds w.h.p. 
\end{theorem}

\vspace{0.1in} \noindent \textbf{Maximum Matching Problems.}
In this section, we study the maximum matching problem. We assume the existence of at least one edge because; otherwise, the problem becomes trivial. We modify a constant time algorithm for the maximum matching problem $\Pi_{\text{MM}}$ and prove that rGINs can simulate this algorithm. Nguyen et al. \cite{nguyen2008constant} converted a sequential greedy algorithm into a constant time algorithm. We use a slightly different version of the sequential algorithm. This algorithm constructs the solution $M$ from the empty set by the following procedure.
\vspace{-0.05in}
\begin{enumerate}
    \item Assign a random value $\boldr_v \sim \mu$ to each node $v \in V$
    \item Let $F = \{\{u, v\} \in E \mid \exists s ,t \in \mathcal{N}_t(u) \cup \mathcal{N}_t(v) \text{ s.t. } \boldr_s = \boldr_t \}$
    \item For $k = 1, 2, \dots t$, \begin{enumerate}
        \item $A \leftarrow \varnothing$
        \item for all paths $P = (v_1, v_2, \dots, v_{k+1})$ of length $k$ in the lexicographical order of $(\boldr_{v_1}, \dots, \boldr_{v_{k+1}})$, if $P$ contains no edge in $F$, is an augment path of $M$, and contains no vertex that appears in $A$, then add all edges of $P$ into $A$
        \item $M \leftarrow M \oplus A$ (i.e., symmetric difference of sets)
    \end{enumerate}
\end{enumerate}
\vspace{-0.05in}

\begin{lemma} \label{lem: MM_sequential}
For all $\varepsilon > 0$, there exist $t \in \mathbb{Z}^+$ and $p \in \mathbb{R}^+$ such that for any distribution $\mu \in \mathcal{U}(p)$, the algorithm is a consistent $(1 + \varepsilon)$-approximation algorithm for the maximum matching problem.
\end{lemma}

Similar to the minimum dominating set problem, this sequential algorithm can be converted to a constant time algorithm and simulated by rGINs.

\begin{theorem} \label{thm: MM}
For all $\varepsilon > 0$, there exist parameters $\boldtheta$, $p \in \mathbb{R}^+$ such that for all distributions $\mu \in \mathcal{U}(p)$ and graphs $G \in \mathcal{F}(\Delta)$, $\textup{rGIN}_E(G, \mu, \boldtheta) \in \Pi_{\text{MM}}(G)$ always holds and $|\textup{rGIN}_E(G, \mu, \boldtheta)| \ge \frac{1}{1 + \varepsilon} ~\textup{OPT}_M(\Pi_{\text{MM}}, G)$ holds w.h.p. 
\end{theorem}

\vspace{0.1in} \noindent \textbf{Opposite Direction.}
So far, we have demonstrated that a certain type of constant algorithm can be converted to randomized GNNs. Next, we prove the opposite direction. The next theorem indicates that the advancement of GNN theory promotes the theory of constant time algorithms.

\begin{theorem} \label{thm: opposite}
If there exist parameters of rGINs that represent a consistent minimization (resp. maximization) algorithm of $\Pi$ with approximation ratio $\alpha$, there exists a constant time algorithm that estimates an $(\alpha, \varepsilon n)$-approximation of $\textup{OPT}_m(\Pi, G)$ (resp. $\textup{OPT}_M(\Pi, G)$) w.h.p.
\end{theorem}

\begin{table*}[tb]
\vspace{-0.1in}
\small
    \caption{Each value stands for an ROC-AUC score. $^*$ denotes a statistically significant improvement for the Wilcoxon signed rank test with $\alpha = 0.05$.} 
    \centering
    \scalebox{0.8}{
    \begin{tabular}{lcccccccll} \toprule
    & TRI(N) & TRI(X) & LCC(N) & LCC(X) & MDS(N) & MDS(X) & MUTAG & \multicolumn{1}{c}{NCI1} & \multicolumn{1}{c}{PROTEINS} \\ \midrule
    GINs & 0.500 & 0.500 & 0.500 & 0.500 & 0.500 & 0.500 & \textbf{0.946} $\pm$ \textbf{0.034} & 0.870 $\pm$ 0.009 & \textbf{0.806} $\pm$ \textbf{0.029} \\
    rGINs & \textbf{0.908} & \textbf{0.926} & \textbf{0.811} & \textbf{0.852} & \textbf{0.807} & \textbf{0.810} & \textbf{0.949} $\pm$ \textbf{0.040} & \textbf{0.876} $\pm$ \textbf{0.010} $^*$ & \textbf{0.810} $\pm$ \textbf{0.030} \\ \midrule
    GCNs & 0.500 & 0.500 & 0.500 & 0.500 & 0.500 & 0.500 & \textbf{0.890} $\pm$ \textbf{0.092} & \textbf{0.819} $\pm$ \textbf{0.023} & 0.804 $\pm$ 0.025 \\
    rGCNs & \textbf{0.855} & \textbf{0.877} & \textbf{0.784} & \textbf{0.785} & \textbf{0.769} & \textbf{0.769} & \textbf{0.904} $\pm$ \textbf{0.099} & \textbf{0.816} $\pm$ \textbf{0.016} & \textbf{0.812} $\pm$ \textbf{0.029} $^*$ \\ \bottomrule
    \end{tabular}
    }
    \label{table: experiments}
    \vspace{-0.2in}
\end{table*}

\section{Experiments} \label{sec: experiments}

We confirm the theoretical results via numerical experiments. We slightly modify the original implementation of GINs to introduce random features. We also use graph convolutional networks (GCNs) and GCNs with random features (rGCNs) to confirm that the addition of the random features can improve the expressive power of GNN architectures other than GINs. In the experiments, we use the uniform distribution over $D = \{0, 0.01, 0.02, \dots, 0.99\}$ as the random distribution $\mu$. The experimental setup and implementation details have been described in the supplementary material.

\vspace{0.1in} \noindent \textbf{Learning Substructures.}
We confirm that rGINs can distinguish local substructures, proven in Theorem \ref{thm: capability}. We use four synthetic datasets in this experiment. 

\begin{itemize}
    \item TRIANGLE: This dataset contains random $3$-regular graphs for a binary node classification problem. Both training and test data contain $1000$ graphs. The training graphs have $20$ nodes, and test graphs have $20$ nodes for the normal dataset (denoted by (N)) and $100$ nodes for the extrapolation dataset (denoted by (X)). A node $v$ is positive if $v$ has two neighboring nodes that are adjacent to each other.
    \item LCC: This dataset contains random $3$-regular graphs for a multi-label node classification problem. Both training and test data contain $1000$ graphs. The training graphs have $20$ nodes, and test graphs have $20$ nodes for the normal dataset and $100$ nodes for the extrapolation dataset. The class of node $v$ is the local clustering coefficient \cite{watts1998collective} of $v$.
\end{itemize}

Learning local clustering coefficients is important because this is a useful feature for spam detection and estimating content quality \cite{becchetti2008efficient, welser2007visualizing}. The test graphs of TRIANGLE(X) and LCC(X) have more nodes than the training graphs. As stated in Sections \ref{sec: introduction} and \ref{sec: main}, the advantage of rGINs is that they can generalize to graphs with variable size. We confirm this using the extrapolation datasets. These datasets are available in \url{https://github.com/joisino/random-features}.

We measure the ROC-AUC scores for TRIANGLE(N) and TRIANGLE(X), and measure the average of the AUC score for each category of the LCC(N) and LLC(X) datasets because they are multi-label problems. We train the models using the cross entropy loss and stochastic gradient descent. The first four columns of Table \ref{table: experiments} report the AUC scores for the test data of these datasets. This indicates that rGINs and rGCNs can learn substructures from data whereas GINs and GCNs cannot distinguish substructures in these datasets. Indeed, the existence of a triangle or the local clustering coefficient can be added as a node feature by hand. However, it is important to note that rGINs and rGCNs can learn these structures from the data without including these structures as node features explicitly. This indicates that rGINs and rGCNs can implicitly utilize the characteristic substructures for positive or negative examples (e.g., chemical functional groups) for classification. This is desirable because there are too many substructures that can affect the performance to include all substructures into node features.

\begin{figure}
    \centering
    \includegraphics[width=0.8\hsize]{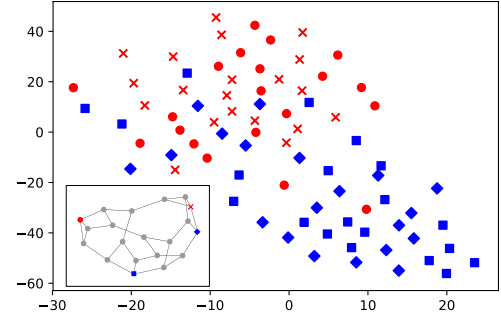}
    \vspace{-0.1in}
    \caption{Scatter plot of node embeddings generated by rGINs with different random feature seeds.}
    \label{fig: triangle_plot}
\end{figure}

As explained in the intuitive explanation, rGINs can classify nodes correctly even if the random features are changed. We confirm this fact via visualization. We sample a test graph $G$ from TRIANGLE(N) and calculate the embedding of two positive nodes and two negative nodes using the trained rGINs in the TRIANGLE(N) experiment with $20$ different random seeds. Figure \ref{fig: triangle_plot} visualizes the test graph $G$ and the scatter plot of the two-dimensional t-SNE embedding \cite{tSNE}. Red (circle and cross) nodes are positive examples, which have a triangle, and blue (square and diamond) nodes are negative examples. This figure shows that rGINs can correctly embed nodes even if the random features change. Note that Figure \ref{fig: triangle_plot} is not completely linear separable because this projects the original $64$-dimensional embeddings into the $2$-dimensional space, and the learned parameters of rGINs are suboptimal.

\vspace{0.1in} \noindent \textbf{Learning Algorithms.}
We confirm that rGINs can learn algorithms for combinatorial problems. In particular, we confirm that rGINs can learn the sequential greedy algorithm \cite{johnson1974approximation, lovasz1975ratio} for the minimum dominating set problem, as shown in Theorem \ref{thm: MDS}. In this experiment, the training graphs and test graphs are random $3$-regular graphs. Both the training and test data contain $1000$ graphs. The training graphs have $20$ nodes and the test graphs of MDS(N) and MDS(X) have $20$ and $100$ nodes, respectively. We set the label of a node as positive if that node is included into the solution of the algorithm and negative otherwise. We train the models using the cross entropy loss and stochastic gradient descent. The fifth and sixth columns of Table \ref{table: experiments} report the AUC scores for the test data of these datasets. It shows that rGINs and rGCNs can learn some concept of the sequential algorithm whereas GINs and GCNs cannot.

\vspace{0.1in} \noindent \textbf{Real World Datasets.}
We confirm the effect of random features on real world datasets: MUTAG, NCI1, and PROTEINS. We measure the cross validation scores following GINs \cite{GIN} and PSCN \cite{PatchySAN} because the evaluation is unstable due to the small dataset sizes. We use this evaluation protocol to show the improvement over GINs. It should be noted that the hyperparameters to be tuned in the cross validation are the same for GINs and rGINs because the distribution of random features was fixed beforehand. The sixth to eighth columns of Table \ref{table: experiments} summarize the AUC scores, which show that rGINs are comparable or slightly outperform the ordinary GINs. In particular, rGINs and rGCNs gain statistically significant improvements in two tasks. The performance gain is less drastic compared to TRIANGLE and LCC datasets because the degrees of nodes differ in the real world datasets, and the original GINs can distinguish most nodes using the degree signals. However, this experiment shows that adding random features does not harm the performance for real world datasets. For example, GINs cannot distinguish between the Decalin \ce{C10H18} and the Bicyclopentyl \ce{C10H18} \cite{sato2020survey}, which means that GINs always fail to classify these molecules if they belong to different categories. In contrast, rGINs can distinguish them by the existence of cycles of length five. This result suggests that the addition of the random features is a handy practice to ensure the capability of GNNs without harming the performance even with slight improvements.

\section{Conclusion}

In this paper, we show that the addition of the random features theoretically strengthens the capability of GINs. Especially, GINs with random features (rGINs) can distinguish any local substructures w.h.p. and solve the minimum dominating set problem and maximum matching problem with nearly optimal approximation ratios. The main advantage of rGINs is that they can guarantee the capability even with arbitrarily large test graphs. In the experiments, we show that rGINs can solve three problems that the normal GINs cannot solve, i.e., determining the existence of a triangle, computing local cluster coefficients, and learning an algorithm for the minimum dominating set problem. We also show that rGINs slightly outperform the normal GINs in biological real world datasets.

\section*{Acknowledgments}

This work was supported by the JSPS KAKENHI GrantNumber 20H04243 and 20H04244.

\bibliographystyle{siam}
\bibliography{citation}

\clearpage

\appendix

\section{Experimental Setups}

\subsection{Dataset Descriptions}

\begin{itemize}
\item \textbf{MUATG} is a chemical compound dataset. Each node represents an atom and each edge represents a chemical bond. A node feature vector is a one-hot vector that represents an atom type. Graphs are labelled according to their mutagenic effect on a bacterium.
\item \textbf{NCI1} is a chemical compound dataset. Each node represents an atom and each edge represents a chemical bond. A node feature vector is a one-hot vector that represents an atom type. Graphs are labelled according to the bioassay records for anti-cancer screen tests with non-small cell lung.
\item \textbf{PROTEINS} is a protein dataset. Each nodes represents a secondary structure element, and a node feature vector is a one-hot vector that represents a structure type (helix, sheet, or turn). Two nodes are connected if they are neighbors either in the amid acid sequence or in 3D space. Graphs are labelled according to whether they are enzymes or non-enzymes.

\end{itemize}

\subsection{General Setups}

In the experiments, we use five-layered GNNs (including the input layer). In our experiments, the graph convolutional networks (GCNs) aggregate the features by average pooling, following \cite{GIN} whereas the original model \cite{GCN} uses symmetrized normalization. We train models with the Adam optimizer \cite{adam} with an initial learning rate of $0.01$ and batch size of $32$. We decay the learning rate by $0.5$ every $50$ epochs. We use dropout in the final layer with a dropout rate of $0.5$ for graph classification datasets (i.e., MUTAG, NCI1, and PROTEINS). We do not train the parameters $\varepsilon$ because the authors of \cite{GIN} showed that it does not affect the performance. We train models for $350$ epochs for the TRIANGLE and LCC datasets, and select the number of epochs from $\{1, 2, \dots, 350\}$ by cross-validation for MUTAG, NCI1, and PROTEINS datasets. It is noteworthy that these hyperparamter settings are the default values of the open implementation of GINs \url{https://github.com/weihua916/powerful-gnns}, which have been shown to be effective in practice.

\vspace{0.1in} \noindent \textbf{Setups for MDS.} We use different hyperparameters in the MDS experiments because learning the greedy algorithm requires more expressive power. Specifically, we use ten-layered GNNs with $1024$ hidden dimensions. We train the model with SGD with learning rate $0.1$ for $50000$ epochs. 

\subsection{Graph Synthesis Process}

\begin{figure*}[tb]
\begin{minipage}{0.24\hsize}
\begin{center}
\includegraphics[width=\hsize]{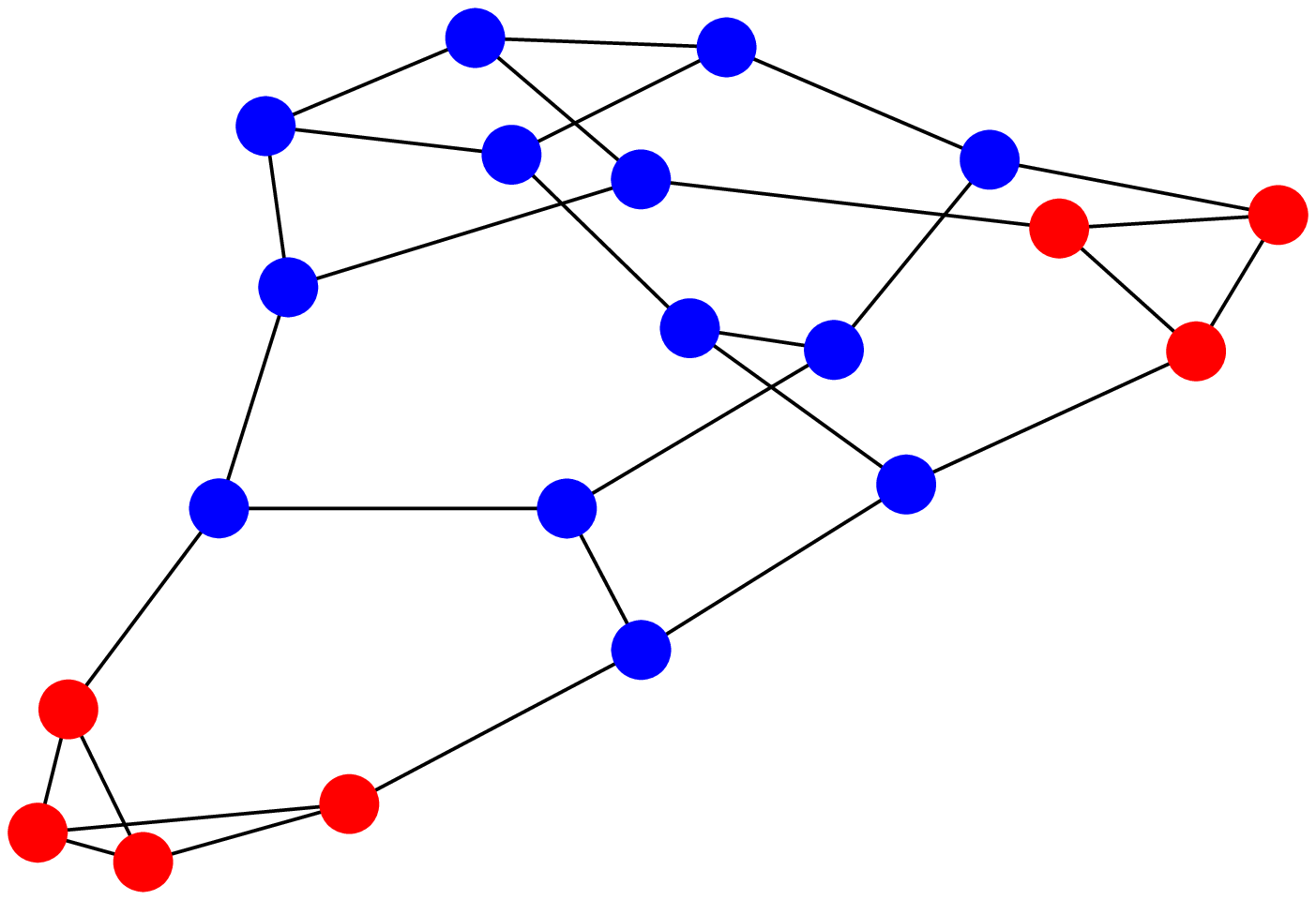}
\end{center}
\end{minipage}
\begin{minipage}{0.24\hsize}
\begin{center}
\includegraphics[width=\hsize]{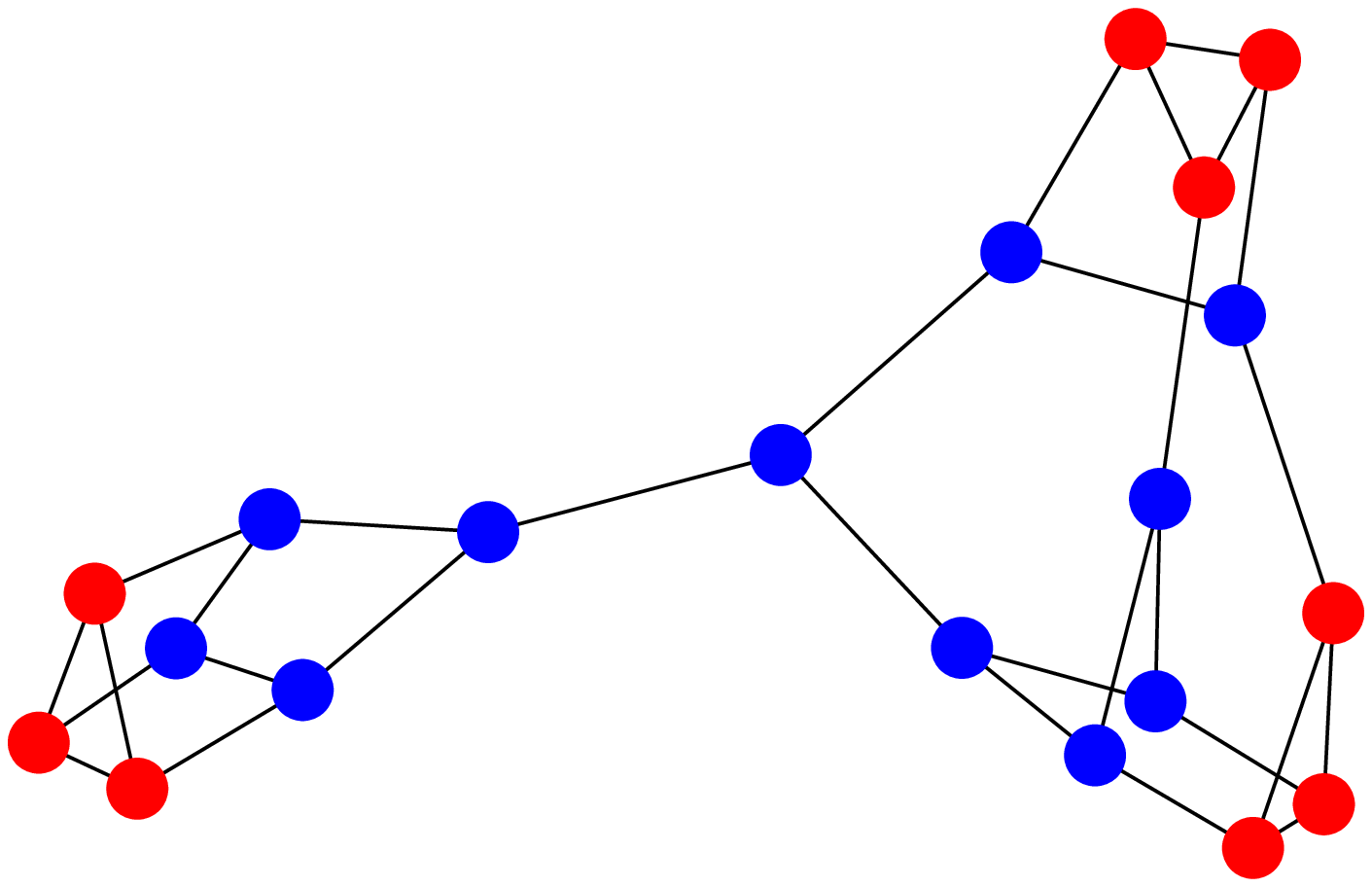}
\end{center}
\end{minipage}
\begin{minipage}{0.24\hsize}
\begin{center}
\includegraphics[width=\hsize]{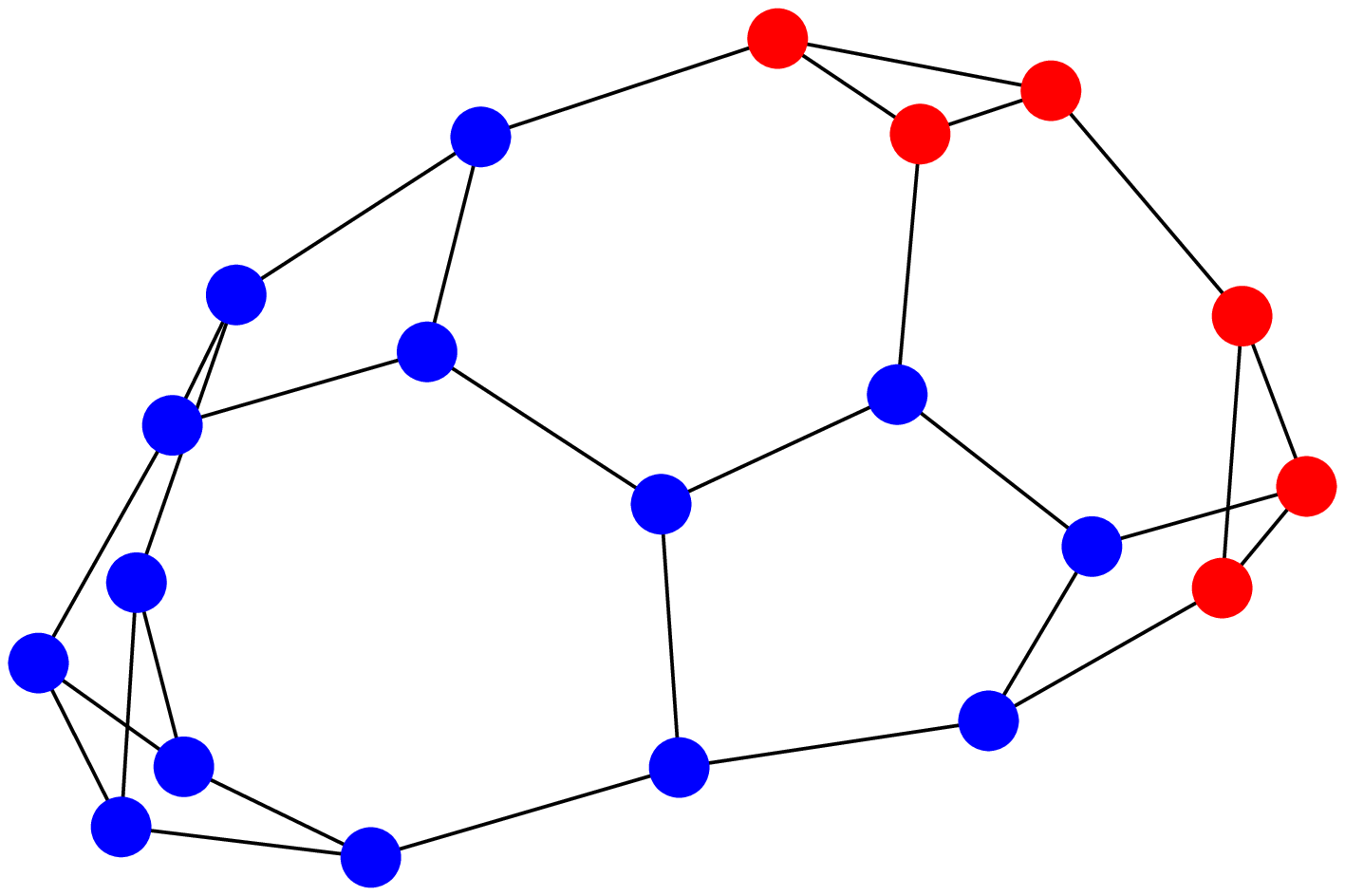}
\end{center}
\end{minipage}
\begin{minipage}{0.24\hsize}
\begin{center}
\includegraphics[width=\hsize]{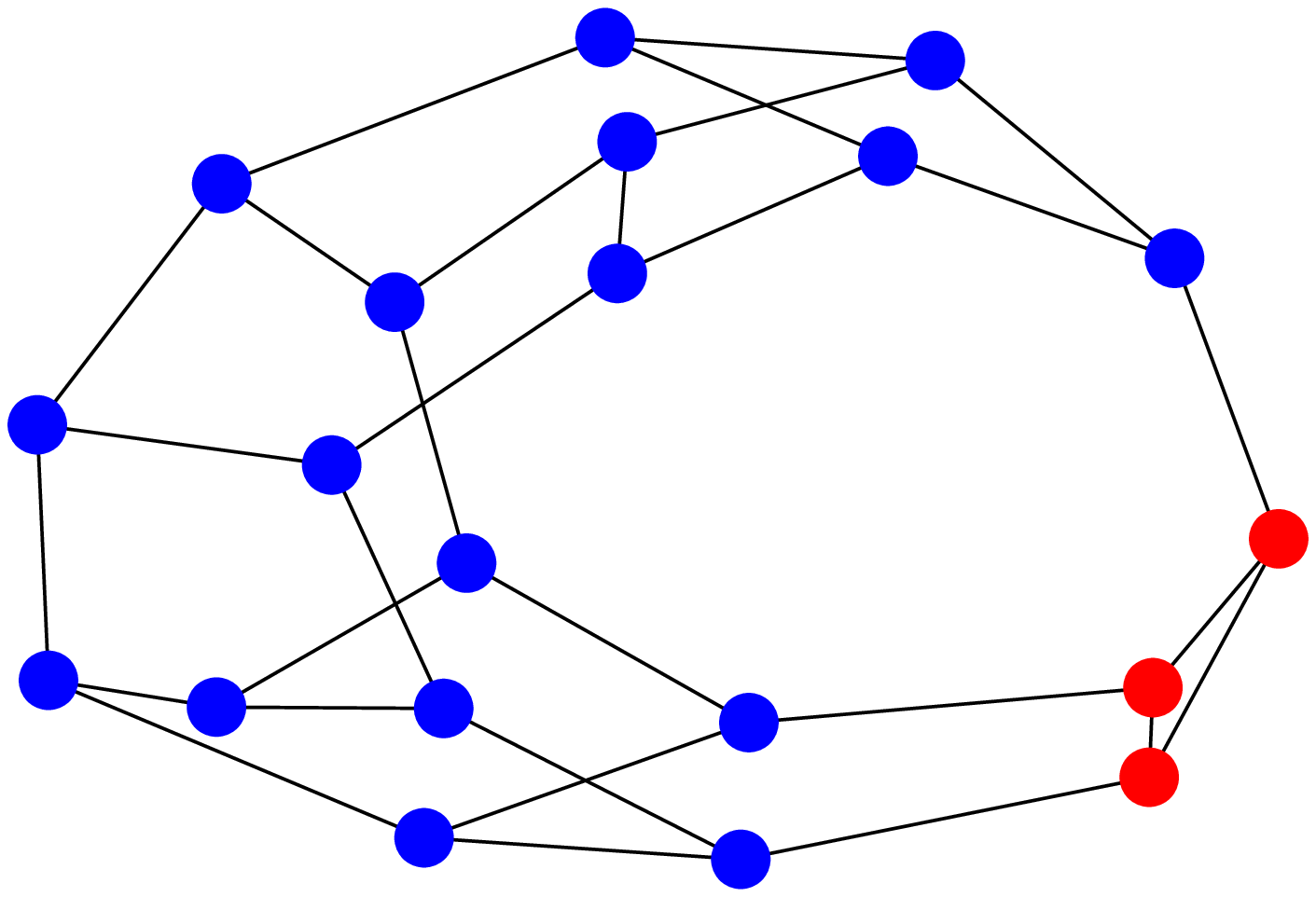}
\end{center}
\end{minipage}
\caption{Examples of the TRIANGLE(S) dataset.} 
\label{fig: tri_example}
\end{figure*}

We generate graphs for TRIANGLE, LCC, and MDS datasets by the \texttt{random\_degree\_sequence\_graph} function of \texttt{networkx} package \cite{bayati2010sequential, NetworkX}. All graphs are generated by the same process with different seeds. Figure \ref{fig: tri_example} shows examples of test graphs of the TRIANGLE(S) dataset.

\section{More Discussion on Assumptions}

We clarify the assumptions of theoretical results in detail. First, the number of dimensions is fixed but might be large, and we assume that the multi-layer perceptron is universal. These assumptions are stated in ``there exist parameters $\boldtheta$'' in the statements of theorems. The universal approximation theorem \cite{cybenko1989approximation, hornik1989multilayer, hornik1991approximation} justifies the latter assumption. Secondly, $p$ depends on the size of $R(G, v, L)$, but does not depend on the size of input graphs. Note that the size of $R(G, v, L)$ is bounded if $L$ and the maximum degree $\Delta$ are bounded.

\section{Proofs}

\begin{lemma} \label{lem: no_local_same_color}
$\forall L \in \mathbb{Z}^+, \varepsilon > 0, \exists p > 0$ s.t. $\forall \mu \in \mathcal{U}(p), \forall G = (V, E) \in \mathcal{F}(\Delta)$, let $\boldr_v \sim \mu$, then $\forall v \in V, Pr[\exists x, y \in \mathcal{N}_L(v), \boldr_x = \boldr_y] < \varepsilon$. 
\end{lemma}

\newproofln{@no_local_same_color}{\ref{lem: no_local_same_color}}
\begin{@no_local_same_color}
Let $L \in \mathbb{Z}^+$ be an arbitrary positive integer and $\varepsilon > 0$ be an arbitrary positive number. Let $p = \Delta^{-2(L+2)} \varepsilon$. $\forall \mu \in \mathcal{U}(p), \forall G = (V, E) \in \mathcal{F}(\Delta) \forall s, t \in V$, $Pr[\boldr_s = \boldr_t] \le p$. Because $|\mathcal{N}_L(v)| < \Delta^{L+2}$, 
\[ \forall v \in V, Pr[\exists x, y \in \mathcal{N}_L(v), \boldr_x = \boldr_y] < \Delta^{2(L+2)} p = \varepsilon. \]
\end{@no_local_same_color}

\noindent \textbf{Definition ($T$):} For $G \in \mathcal{F}(\Delta, C)$, $\boldR \in D^n$, $v \in G$, and $l \in \mathbb{Z}^+$, let $T(G, v, l)$ be
\begin{align*}
    &T(G, v, 0) = (\boldx_v, \boldr_v) \\
    &T(G, v, l) \\ &= (T(G, v, l-1), \textsc{Multiset}( T(G, u, l-1) \mid u \in \mathcal{N}(v) ))
\end{align*}
We call $T(G, v, l)$ a level-$l$ tree.

\begin{corollary}[from \cite{GIN}] \label{cor: GIN_injective}
For all $l \in \mathbb{Z}^+$, there exist parameters $\boldtheta$ of GINs such that for all $G = (V, E, \boldX), G' = (V', E', \boldX') \in \mathcal{F}(\Delta, C)$, $\boldR \in D^{|V|}$, $\boldR' \in D^{|V'|}$, $v \in V$, $v' \in V'$ if $T(G, v, L) \neq T(G', v', L)$ holds, $\textup{GIN}_\boldtheta((V, E, [\boldX, \boldR]))_v \neq \textup{GIN}_\boldtheta((V', E', [\boldX', \boldR']))_{v'}$ holds.
\end{corollary}

\begin{lemma} \label{lem: graph_from_tree}
For all $L \in \mathbb{Z}^+$, there exists a function $f$ such that for all $G = (V, E, \boldX) \in \mathcal{F}(\Delta, C)$, $\boldR \in D^n$, $v \in G$, if for all $s, t \in \mathcal{N}_{L+1}(v)$, $\boldr_s \neq \boldr_t$ holds, 
\[ f(T(R(G, v ,L), v, L + 1)) \simeq (R(G, v, L), v). \]
\end{lemma}

\newproofln{@graph_from_tree}{\ref{lem: graph_from_tree}}
\begin{@graph_from_tree}
We construct $f(T) \in \mathcal{F}(\Delta)$. Let the node set $V'$ be all $\boldr_p$ of level-$0$ sets $[\boldx_p, \boldr_p]$ that $T[0]$ (i.e., the left element of $T$) contains, and the feature vector $\boldx'_{\boldr_p}$ be equal to one of $\boldx_p$ (e.g., the smallest one). If for all $s, t \in \mathcal{N}_L(v)$, $\boldr_s \neq \boldr_t$ holds, the choice of $\boldx'_{\boldr_p}$ is unique, and the number of nodes is the same as $G$. Let the center node be $v' = T[0][0] \dots [0][1]$ (i.e., the right element of the leftmost level-$0$ set). $v'$ is equal to $\boldr_v$ by construction. For $p, q \in V'$, $\{p, q\}$ is included in $E'$ if and only if there exists a level-$1$ set $T'$ such that the right element of the left element of $T'$ is equal to $p$, and a level-$0$ set whose right element is equal to $q$ is included in the right element of $T'$. There exists an edge between $\{s, t\} \in E$ if and only if there exists an edge between $\{\boldr_s, \boldr_t\} \in E'$ by construction. Therefore, $f(T) = ((V', E', \boldX'), v') \simeq (R(G, v, L), v)$ holds. 
\end{@graph_from_tree}

\newprooftn{@capability}{\ref{thm: capability}}
\begin{@capability}
From Corollary \ref{cor: GIN_injective} and Lemma \ref{lem: graph_from_tree}, for all $L \in \mathbb{Z}^+$, there exist parameters $\boldtheta$ of GINs such that for all $G = (V, E, \boldX), G' = (V', E', \boldX') \in \mathcal{F}(\Delta, C)$, $\boldR \in D^{|V|}$, $\boldR' \in D^{|V'|}$, $v \in V$, $v' \in V'$ if (1) for all $s, t \in \mathcal{N}_{L+1}(v)$, $\boldr_s \neq \boldr_t$, (2) for all $s', t' \in \mathcal{N}_{L+1}(v')$, $\boldr_{s'} \neq \boldr_{t'}$, and (3) $(R(G, v, L), v) \not \simeq (R(G', v', L), v')$ hold, $\textup{GIN}_\boldtheta((V, E, [\boldX, \boldR]))_v \neq \textup{GIN}_\boldtheta((V', E', [\boldX', \boldR']))_{v'}$ holds. The cardinality of $\{R((V, E, [\boldX, \boldR]), v, L) \mid G = (V, E, \boldX) \in \mathcal{F}(\Delta, C), v \in V, \boldR \in D^{|V|}\}$ is finite. Therefore, there exists a multi layer perceptron MLP such that if the first and second conditions are valid, $\textup{MLP}(\textup{GIN}_\boldtheta((V, E, [\boldX, \boldR]))_v) > 0.5$ if $\exists (G', v') \in \mathcal{G}$ such that $(G', v') \simeq (R(G, v, L), v)$ holds and $\textup{MLP}(\textup{GIN}_\boldtheta((V, E, [\boldX, \boldR]))_v) < 0.5$ otherwise. This MLP can be considered as the last layer of the GIN. From Lemma \ref{lem: no_local_same_color}, there exists $p$ such that for all $\mu \in \mathcal{U}(p)$, the probability of the first and second conditions being true is arbitraily small. 
\end{@capability}

\newproofln{@MDS_sequential}{\ref{lem: MDS_sequential}}
\begin{@MDS_sequential}{\ref{lem: MDS_sequential}}
Let $U = \{v \in V \mid \exists u \in \mathcal{N}_2(v) \text{ s.t. } \boldr_v = \boldr_u \}$ and $U^+ = \{v \in V \mid \exists u \in \mathcal{N}(v) \text{ s.t. } u \in U \}$. The solution $S$ our sequential algorithm outputs for $G$ is the same as the union of $U^+$ and the solution $T$ the sequential algorithm \cite{nguyen2008constant, lovasz1975ratio, johnson1974approximation} outputs for the induced graph $G'$ of $V \backslash U^+$ by construction. From Lemma \ref{lem: no_local_same_color}, there exists $p$ such that for all $\mu \in \mathcal{U}(p)$, $|U| \le \frac{\varepsilon}{\Delta + 1} n$ with high probability. It means that $|U| \le \varepsilon \textup{OPT}_m(\Pi_{\text{MDS}}, G')$ because for any $G = (V, E) \in \mathcal{F}(\Delta)$, $\textup{OPT}_m(\Pi_{\text{MDS}}, G) \ge \frac{n}{\Delta + 1}$ holds. Therefore, with high probability,
\begin{align*}
    |S|
    &\le |U| + |T| \\
    &\le  (H(\Delta + 1) + \varepsilon) ~ \textup{OPT}_m(\Pi_{\text{MDS}}, G') \\
    &\le  (H(\Delta + 1) + \varepsilon) ~ \textup{OPT}_m(\Pi_{\text{MDS}}, G)
\end{align*}
\end{@MDS_sequential}

\newproofln{@MDS_local}{\ref{lem: MDS_local}}
\begin{@MDS_local}
Let $S_L = \{ v \in V \mid \text{the computation of } \mathcal{O}_{\Delta + 1}(v) \text{ stops within } R(G, v, L) \}$, $U_L = \{ v \in V \mid \exists s, t \in \mathcal{N}_L(v) \text{ s.t. } \boldr_s = \boldr_t \}$, $A = \{ v \in V \mid \mathcal{O}_{\Delta + 1}(v) = 1 \}$. For fixed $\boldR$, let $f$ be a function that outputs $1$ if $v \in F = S_L \cup U_L \cup A$, and $0$ otherwise. This value can be computed only from $(R(G, v, L), v)$. From the locality lemma \cite{nguyen2008constant}, there exists $L \in \mathbb{Z}^+$ such that $|S_L| \le \frac{\varepsilon}{3(\Delta + 1)} n \le \frac{\varepsilon}{3} ~\textup{OPT}_m(\Pi_{\text{MDS}}, G)$ holds w.h.p. From Lemma \ref{lem: no_local_same_color}, $|U_L| \le \frac{\varepsilon}{3(\Delta + 1)} n \le \frac{\varepsilon}{3} ~\textup{OPT}_m(\Pi_{\text{MDS}}, G)$ holds w.h.p. if $p$ is sufficiently large. From Lemma \ref{lem: MDS_sequential}, there exists $p$ such that for all $\mu \in \mathcal{U}(p)$, $|A| \le (H(\Delta + 1) + \frac{\varepsilon}{3}) ~\textup{OPT}_m(\Pi_{\text{MDS}}, G)$ holds w.h.p. $F$ always forms a dominating set because $A \subseteq F$. From the above equations, $|F| \le |S_L| + |U_L| + |A| \le (H(\Delta + 1) + \varepsilon) ~\textup{OPT}_m(\Pi_{\text{MDS}}, G)$ holds w.h.p.
\end{@MDS_local}

\newprooftn{@MDS}{\ref{thm: MDS}}
\begin{@MDS}
From Corollary \ref{cor: GIN_injective}, Lemma \ref{lem: graph_from_tree}, and Lemma \ref{lem: MDS_local}, there exist parameters $\boldtheta$ of GINs such that $\forall G = (V, E) \in \mathcal{F}(\Delta)$, $\forall \boldR \in D^{|V|}$,
\begin{align*}
 &\{v \in V \mid \text{GIN}_\boldtheta((V, E, \boldR))_v > 0.5\} \\
 &\quad \cup \{ v \in V \mid \exists s, t \in \mathcal{N}_L(v) \text{ s.t. } \boldr_s = \boldr_t \} \\
 &= \{ v \in V \mid f(R((V, E, \boldR), v, L), v) = 1 \},
\end{align*}
where $f$ is the function in the proof of Lemma \ref{lem: MDS_local}. Therefore, rGINs can simulate $f$, which is $(H(\Delta + 1) + \varepsilon)$-approximation w.h.p.
\end{@MDS}

\newproofln{@MM_sequential}{\ref{lem: MM_sequential}}
\begin{@MM_sequential}
Let $F = \{e = \{u, v\} \in E \mid \exists s, t \in \mathcal{N}_t(u) \cup \mathcal{N}_t(v) \text{ s.t. } \boldr_s = \boldr_t \}$. The solution $S$ our sequential algorithm outputs for $G$ is the same as the solution $T$ the sequential algorithm \cite{nguyen2008constant} outputs for $G' = (V, E \backslash F)$. Let $\varepsilon' = \frac{\varepsilon}{2}$. From \cite{nguyen2008constant}, the sequential algorithm \cite{nguyen2008constant} is $(1 + \varepsilon')$-approximation by setting $t = \text{ceil}(\frac{1}{\varepsilon'})$. From Lemma \ref{lem: no_local_same_color}, there exists $p \in \mathbb{R}^+$ such that for all $\mu \in \mathcal{U}(p)$, $|F| \le \frac{\varepsilon}{8 (1 + \varepsilon)^2 \Delta} n$ because $|E| \le \frac{\Delta}{2} n$ by the degree-bounded assumption. When an edge is added to the solution, at most $2 \Delta$ candidate edges are excluded. Therefore, 
\[ \text{OPT}_M(\Pi_{\text{MM}}, G) \ge \frac{m}{2 \Delta} \ge \frac{n - 1}{2 \Delta} \ge \frac{n}{4 \Delta} \]
holds. Furthermore,
\[ \frac{1}{1 + \varepsilon'} - \frac{1}{1 + \varepsilon} = \frac{\varepsilon - \varepsilon'}{(1 + \varepsilon')(1 + \varepsilon)} \ge \frac{\varepsilon - \varepsilon'}{(1 + \varepsilon)^2} = \frac{\varepsilon}{2(1 + \varepsilon)^2} \]
holds. Therefore,
\begin{align*}
    |S| = |T| &\ge \frac{1}{1 + \varepsilon'} \text{OPT}_M(\Pi_{\text{MM}}, G') \\
    &\ge \frac{1}{1 + \varepsilon'} (\text{OPT}_M(\Pi_{\text{MM}}, G) - |F|) \\
    &\ge \frac{1}{1 + \varepsilon'} \text{OPT}_M(\Pi_{\text{MM}}, G) - |F| \\
    &\ge \frac{1}{1 + \varepsilon'} \text{OPT}_M(\Pi_{\text{MM}}, G) - \frac{\varepsilon}{8 (1 + \varepsilon)^2 \Delta} n \\
    &\ge \frac{1}{1 + \varepsilon'} \text{OPT}_M(\Pi_{\text{MM}}, G) \\ & \qquad \qquad - \frac{\varepsilon}{2 (1 + \varepsilon)^2}\text{OPT}_M(\Pi_{\text{MM}}, G) \\
    &\ge \frac{1}{1 + \varepsilon} \text{OPT}_M(\Pi_{\text{MM}}, G).
\end{align*}
This concludes that our sequential algorithm is $(1 + \varepsilon)$-approximation.
\end{@MM_sequential}

\newprooftn{@thmMM}{\ref{thm: MM}}
\begin{@thmMM}
We assume $D = [k]$ for some $k \in \mathbb{Z}^+$ without loss of generality. Let $S$ be the solution that our algorithm outputs for $G = (V, E)$. By the locality lemma and Lemma 7 of \cite{nguyen2008constant}, there exists $L \in \mathbb{Z}^+$ such that there exists an algorithm $\mathcal{A}$ that takes $G' = (V, E, \boldR)$ and $e = \{v, u\} \in E$ and decides $e \in S$ or not with the following property: Let $T = \{ e \in E \mid \mathcal{A} \text{ accesses } G \backslash (R(G', u, L) \cap R(G', v, L)) \text{ to decide } e \in S \text{ or not } \}$, then $|T| \le \frac{\varepsilon}{16 \Delta} n$ w.h.p. with respect to the randomness of $\boldR$. Take $L$ such that $|S| \ge \frac{1}{1 + \varepsilon/4} \text{OPT}_M(\Pi_{\text{MM}}, G)$ holds w.h.p. Let $F = \{e = \{u, v\} \in E \mid \exists s, t \in \mathcal{N}_L(u) \cap \mathcal{N}_L(v) \text{ s.t. } \boldr_s = \boldr_t \}$ and $U = \{v \in V \mid \exists s, t \in \mathcal{N}_L(v) \text{ s.t. } \boldr_s = \boldr_t \}$. Let $g(R(G', v, L), v) \in \mathbb{R}^{k^2}$ be $\mathbf{0} \in \mathbb{R}^{k^2}$ if $v \in U$. Otherwise, $g(R(G', v, L), v)_i = 1$ if there exists an edge $\{u, v\} \in E$ such that $e \in S \backslash (F \cup T)$ and $i = \min(\boldr_v, \boldr_u) k + \max(\boldr_v, \boldr_u) - k$ hold, and $0$ otherwise. $g$ can be computed only from $R(G', v, L)$ and $v$ by construction. For any pair of nodes $u, v \in (V \backslash U)$, $e = \{u, v\} \in S \backslash (F \cup T)$ holds if and only if $g(R(G', u, L), u)^\top g(R(G', v, L), v) = 1$ by construction. Therefore, $A = \{ g(R(G', u, L), u)^\top g(R(G', v, L), v) > 0.5\} \subseteq \Pi_{\text{MM}}(G)$ holds because $A \subseteq S$. By Lemma \ref{lem: no_local_same_color} and Lemma \ref{lem: MM_sequential}, there exists $p$ such that for all $\mu \in \mathcal{U}(p)$, $|A| \ge \frac{1}{1 + \varepsilon} \text{OPT}_M(\Pi_{\text{MM}}, G)$ w.h.p. By the same argument as Theorem \ref{thm: MDS}, $g$ can be simulated by rGINs because $g$ only depends on $R(G', v, L)$, and $g(R(G', v, L), v) = \mathbf{0}$ if $v \in U$.
\end{@thmMM}

\newprooftn{@opposite}{\ref{thm: opposite}}
\begin{@opposite}
Let $\boldtheta$ and $\mu$ be the parameters and distributions of rGINs that represent an $\alpha$ approximation algorithm of $\Pi$. Let $L$ be the number of layers in the rGIN. Draw the random feature $\boldr_u$ when the rGINs first accesses the node $u \in V$. Then, for each $v \in V$, $\text{rGINs}(G, \mu, \boldtheta)_v$ can be computed in a constant time because the size of $R(G, v, L)$ is bounded by a function of $L$ and $\Delta$, which are constant. For a node problem, sample $O(\frac{1}{\varepsilon^2})$ nodes $U \subseteq V$ uniformly randomly, and decide each node $v \in U$ is included in the solution by computing $\text{rGINs}(G, \mu, \boldtheta)_v$. For an edge problem, sample $O(\frac{1}{\varepsilon^2})$ edges $F$ uniformly randomly, and decide each edge $e = \{u, v\} \in F$ is included in the solution by computing the $\text{rGINs}(G, \mu, \boldtheta)_u$ and $\text{rGINs}(G, \mu, \boldtheta)_v$. From the Hoeffding bound, the size of the solution can be estimated with additive error $\varepsilon n$ w.h.p.
\end{@opposite}

\end{document}

%% file: mathdef.tex
\newcommand{\mathbbR}{\mathbb{R}}

\newcommand{\boldR}{{\boldsymbol{R}}}

\newcommand{\boldX}{{\boldsymbol{X}}}

\newcommand{\boldr}{{\boldsymbol{r}}}

\newcommand{\boldv}{{\boldsymbol{v}}}

\newcommand{\boldx}{{\boldsymbol{x}}}

\newcommand{\boldz}{{\boldsymbol{z}}}

\newcommand{\boldtheta}{{\boldsymbol{\theta}}}

\newcommand{\calC}{{\mathcal{C}}}

